\documentclass[10pt,twocolumn,letterpaper]{article}

\usepackage{cvpr}
\usepackage{times}
\usepackage{epsfig}
\usepackage{graphicx}
\usepackage{subfigure}
\usepackage{amsmath}
\usepackage{float}
\usepackage{amssymb}
\usepackage{color}

\usepackage[breaklinks=true,colorlinks,bookmarks=false]{hyperref}

\cvprfinalcopy 

\def\cvprPaperID{****} 

\definecolor{orange}{rgb}{1,0.5,0}
\definecolor{mdgreen}{rgb}{0.05,0.6,0.05}
\definecolor{mdblue}{rgb}{0,0,0.7}
\definecolor{dkblue}{rgb}{0,0,0.5}
\definecolor{dkgray}{rgb}{0.3,0.3,0.3}
\definecolor{slate}{rgb}{0.25,0.25,0.4}
\definecolor{gray}{rgb}{0.5,0.5,0.5}
\definecolor{ltgray}{rgb}{0.7,0.7,0.7}
\definecolor{purple}{rgb}{0.7,0,1.0}
\definecolor{lavender}{rgb}{0.65,0.55,1.0}

\begin{document}

\title{Margin Sample Mining Loss: A Deep Learning Based Method for Person Re-identification}

\author{Qiqi Xiao ,
	    Hao Luo ,
	    Chi Zhang \\
{\tt\small qiqix@andrew.cmu.edu, haoluocsc@zju.edu.cn, zhangchi@megvii.com}
}

\maketitle

\begin{abstract}
Person re-identification (ReID) is an important task in computer vision.
Recently, deep learning with a metric learning loss has become a common framework for ReID.
In this paper, we propose a new metric learning loss with hard sample mining called margin smaple mining loss (MSML) which can achieve better accuracy compared with other metric learning losses, such as triplet loss.
In experiments, our proposed methods outperforms most of the state-of-the-art algorithms on Market1501, MARS, CUHK03 and CUHK-SYSU.
\end{abstract}

\section{Introduction}
\label{introduction}

Person re-identification (ReID) is an important and challenging task in computer vision. 
It has many applications in surveilance video, such as person tracking across multiple cameras and person searching in a large gallery \etc.
However, some issues make the task difficult such as large variations in poses, viewpoints, illumination, background environments and occlusion.
And the similarity of appearances among different persons also increases the difficulty.

Some traditional ReID approaches focus on low-level features such as colors, shapes and local descriptors \cite{farenzena2010person, hamdoun2008person}.
With the development of deep learning, the convolutional neural network (CNN) is commonly used for feature representation \cite{matsukawa2016person, varior2016gated, cheng2016person}.
The CNN based methods can present high-level features and thus improve the performance of person ReID. 
In supervised learning, current methods can be divided into representation learning and metric learning in terms of the target loss.
For the representation learning, ReID is considered as a verification or identification problem. 
For instance, in \cite{zheng2016person}, the authors make the comparison between the verification baseline and the identification baseline:
(1) For the former, two images are judged whether they belong to the same person. 
(2) For the latter, the method treats each identity as a category, and then minimizes the softmax loss.
In some improved work, Lin et al. combined the verification loss with attributes loss \cite{lin2017improving}, while Matsukawa et al. combined the identification loss with attributes loss \cite{matsukawa2016person}. 
Representation learning based methods have prominent advantages, having reasonable performance and being easily trained and reproducible. 
But those methods do not care about the similarity of different pairs, leading it difficult to distinguish between pairs of same persons and different persons. 
To mitigate that problem, different distance losses, such as contrastive loss \cite{varior2016gated}, triplet loss \cite{liu2017end}, improved triplet loss \cite{cheng2016person}, quadruplet loss \cite{chen2017beyond}, \etc are proposed.
And \cite{hermans2017defense} also proposes hard batch by sampling hard image pairs. 
These methods can directly evaluate the similarity of two input images according to their embedding features.
Although these distance losses are sensitive to image pairs, which increases the training difficulty,
they can generally get better performance than representation learning based methods.

\begin{figure}[t]
\centering
\includegraphics[width=.98\linewidth]{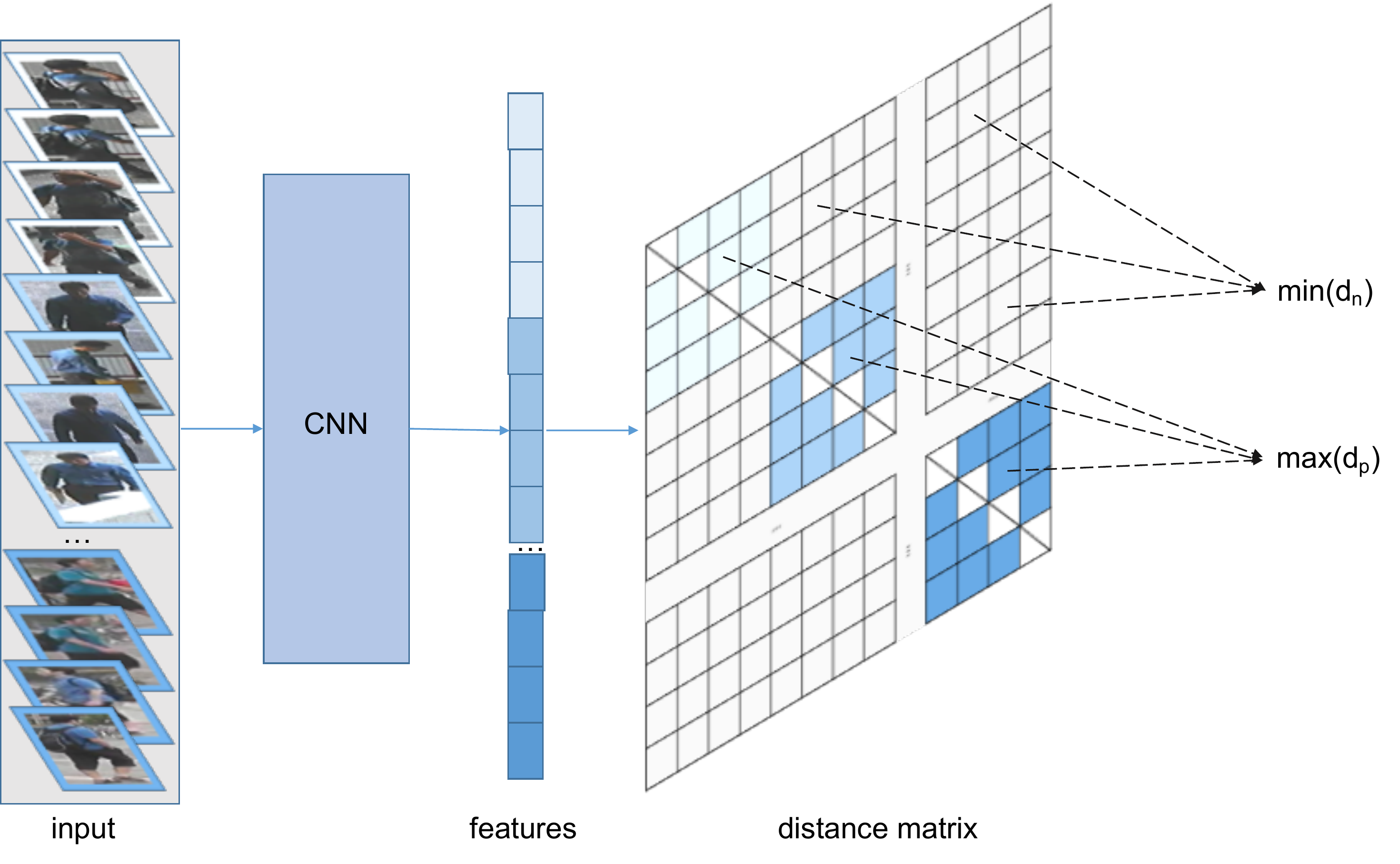}
\caption{Framework of our method. Input data are designed to be groups of identities. Distace matrix of features extracted by CNN is calculated. The minimum of negative pair distances and the maximum of positive pair distances are sent to the loss function.}
\label{fig:framework}
\end{figure}

In this paper, we propose a novel metric learning loss with hard sample mining called margin smaple mining loss (MSML). It can minimize the distance of positive pairs while maximizing the distance of negative pairs in feature embedding space.
For original triplet or quadruplet loss, the pairs are randomly sampled.
In our method, we put each $K$ images of $P$ persons into a batch, and then calculate an $N\times N$ distance matrix where $N=K\times P$ denotes the batch size.
Then, we choose the maximum distance of positive pairs and the minimum distance of negative pairs to calculate the final loss.
In this way, we sample the most dissimilar positive pair and the most similar negative pair, both of which are hardest to be distinguished in the batch.
On Market1501, MARS, CUHK03 and CUHK-SYSU, our method outperforms most of state-of-the-art ones.

In the following, we overview the main contents of our method and summarize the contributions: 
\begin{itemize}
\item{We} propose a new loss with extremely hard sample mining named margin smaple mining loss, which outperforms other metric learning losses on person ReID task.

\item {Our} method shows significant performance on those four datasets, being superior to most of state-of-the-art methods.

\end{itemize}

The paper is organized as follows: related works with more details are presented in section \ref{relatedwork}. 
In section \ref{method}, we introduce our MSML.  
Datasets and experiments are presented in section \ref{experiments}. 
Conclusions and outlook are presented in section \ref{conclusion}.

\section{Related Work}
\label{relatedwork}

\subsection{Deep convolutional networks}
\label{deepnetworks}
Including AlexNet(CaffeNet) \cite{krizhevsky2012imagenet}, GoogleNet \cite{szegedy2015going} and Resnet \cite{he2016deep} etc. , several popular deep networks have been proposed in the past few years. 
A lots of works show that Resnet is better than other baseline models on person ReID task \cite{zhong2017re, zheng2017pose, zheng2017unlabeled}. 
Most current paper choose Resnet50 pre-trained on the ImageNet LSVRC image classification datasets \cite{russakovsky2015imagenet} as baseline networks. 
In this paper, we also choose Resnet50 as our baseline network but reconstruct it. 

Resnet is the origin of deep residual networks, and there are some improved versions such as ResNeXt \cite{xie2016aggregated}, DenseNet \cite{huang2016densely} and ShuffleNet \cite{zhang2017shufflenet} .
All these works use efficient channel-wise convolutions or group convolutions into the building blocks to balance representation capability and computational cost.
Different from traditional regular convolutions, group convolutions divide feature maps into several groups concatenated together after respective convolutions. 
The channel-wise convolutions in which the number of groups equal to the number of channels is a special case of group convolutions. 
Channel-wise convolutions can effectively reduce computational cost. 
GoogLeNet Xception \cite{chollet2016xception} uses a large number of channel-wise convolutions.
Using building blocks designed with group convolutions and channel-wise convolutions to replace regular convolutions in Resnet is popular and improves accuracy with less computational cost. 
\subsection{Deep metric learning}
\label{deepmetriclearning}
Before deep learning, most traditional metric learning methods focus on learning a Mahalanobis distance in Euclidean space. 
Cross-view Quadratic Discriminant Analysis (XQDA) \cite{liao2015person} and  Keep It Simple and Straightforward Metric Learning (KISSME) \cite{koestinger2012large} were both classic metric learning methods in person ReID in the past. 
However, deep metric learning methods usually transform raw images into embedding features, and then compute the similarity scores or feature distances directly in Euclidean space.  

In deep metric learning, two images of the same person are defined as a positive pair while two of different persons are a negative pair. 
The triplet loss is motivated by the threshold enforced between positive and negative pairs. 
In improved triplet loss, a distance loss of positive pairs is used to reinforce clustering of  the same person images in the feature space. 
The positive pair and the negative one in a triplet share a common image. 
A triplet only has two identities.  
Quadruplet loss adds a new negative pair, and a quadruplet samples four images from three identities. 
For the quadruplet loss, a new loss enforces the distance between positive pairs of one identity and negative pairs of the other two identities. 
Deep metric learning methods is sensitive to the samples of pairs. 
Selecting suitable samples for training model by hard mining is shown to be effective \cite{hermans2017defense, chen2017beyond}. 
A common practice is to pick out dissimilar positive pairs and similar negative pairs according to similarity scores. 
Compared with identification or verification loss, distance loss for metric learning can lead to a margin between inter-class distance.
But combining softmax loss with distance loss to speed up convergence is also popular. 
\subsection{Other proposed ReID methods}
Some successful unsupervised or transfer learning methods have been proposed recently \cite{fan2017unsupervised, radenovic2016cnn, peng2016unsupervised, Zhang_2016_CVPR}. 
One important concern is that there exists bias among datasets collected in different environments.
Another problem is the lack of labeled data, which can cause overfitting easily.
Despite that supervised learning methods based on CNN have been successful in some certain dataset, the network trained with that dataset could perform poorly on other datasets. 
There, one method of transfer learning is to train one task with one dataset, and then fine-tune from the trained model to train another task with another dataset.
For example, the model trained on one dataset clusters the other dataset to predict the fake labels which are used to fine-tune the model \cite{fan2017unsupervised}. 
In \cite{peng2016unsupervised}, an unsupervised multi-task dictionary learning is proposed to solve dataset-biased problem. 

In addition, some paper focus on getting better global or local features. 
For instance, pose invariant embedding (PIE) aligns pedestrians to a standard pose to reduce the impact of pose \cite{zheng2017pose} variation. 
Natural language description \cite{li2017person} and image data generated by generative adversarial networks (GANs) \cite{zheng2017unlabeled} are respectively regarded as additional information input into networks. 
Inspite of image-based learning methods above, there are some video-based person ReID works, which take into account the sequence information such as motion or optical flow \cite{wang2016person, zhang2017image, you2016top, ma2017person, mclaughlin2016recurrent, zhao2017person, liu2017video}. 
RNN architectures and attention model  are also applied into embedding sequence features.

After getting image features, most current works choose L2 euclidean distance to compute similarity score for ranking or retrieval task. 
In \cite{wang2017deep, zhong2017re, bai2017scalable}, some re-ranking methods are proposed and obviously improve the ReID accuracy.

\section{Our Method}
\label{method}
Despite the deep network for feature extracting, our method includes a metric learning loss with hard sample mining called MSML.

\subsection{Margin Sample Mining Loss for Metric Learning}

The goal of metric embedding learning is to learn a function $g(x): \mathbb{R}^F \rightarrow \mathbb{R}^D$ which maps semantically similar instances from the data manifold in $\mathbb{R}^F$ onto metrically close points in $\mathbb{R}^D$ \cite{hermans2017defense}.
The deep metric learning aims to find the function through minimizing the metric loss of training data.
Then we should define a metric function $D(x,y): \mathbb{R}^D \times \mathbb{R}^D \rightarrow \mathbb{R}$ to measure the distances in the embedding space.
The distances are used to re-identity the person images.

\subsubsection{Related Metric Learning Methods}
\label{proposedloss}

One of the widely used metric learning loss is triplet loss \cite{schroff2015facenet} which helps generate features as discriminative as possible compared to softmax loss for classification.
It is trained on groups of triplets.
A triplet contains three different images\{$I_A$, $I_{A^\prime}$, $I_B$\}, where $I_A$ and $I_{A^\prime}$ are images of the same identity while $I_B$ is an image of a different identity.
Each image would generate one extracted feature after a deep network.
A triplet of {$\ell_2$-normalized} features \{$f_A$, $f_{A^\prime}$, $f_B$\} would be used to calculate distances and the triplet loss is formulated as following:
\begin{equation}\label{trploss}
L_{trp} = \frac{1}{N}\sum^{N}\left( \overbrace{\left \| f_A - f_{A^\prime} \right \|_{2}}^{\text{to shorten}} - \overbrace{\left \| f_A - f_B \right \|_{2}}^{\text{to largen}} + \alpha \right)_{+}
\end{equation}
where $(z)_+=max(z,0)$ \cite{rosasco2004loss}, and $\alpha$ is the value of the margin set to allow the network distinguish the positive samples with the negative ones.
The first term shortens the distances of positive pairs, while the second term largens the distances of negative pairs.
In triplet loss, each positive pair and negative pair share one same image, which makes it pay more attention to obtaining correct orders for pairs \textit{w.r.t.} the same probe image. 
As a result, it suffers poor generalization, and is difficult to be applied for tracking tasks.

The quadruplet loss \cite{chen2017beyond} extends the triplet loss by adding a different negative pair. 
A quadruplet contains four different images \{$I_A$, $I_{A^\prime}$, $I_B$, $I_C$\}, where $I_A$ and $I_{A^\prime}$ are images of the same identity while $I_B$ and $I_C$ are images of another two identities respectively. 
Accordingly, a quadruplet of {$\ell_2$-normalized} features \{$f_A$, $f_{A^\prime}$, $f_B$, $f_C$\} would be used to calculate distances.
The quadruplet loss is formulated as following:
\begin{equation}\label{quadloss}
\begin{split}
L_{quad} = \frac{1}{N}\sum^N \left( \overbrace{\left \| f_A - f_{A^\prime} \right \|_2 - \left \| f_A - f_B \right \|_2 + \alpha}^{\text{relative distance}} \right)_{+} \\
+ \frac{1}{N}\sum^N \left( \overbrace{\left \| f_A - f_{A^\prime} \right \|_2 - \left \| f_C - f_B \right \|_2 + \beta}^{\text{absolute distance}} \right)_{+}
\end{split}
\end{equation}
where $\alpha$ and $\beta$ are the values of the margins in two terms.
The first term is the same as \eqref{trploss}, which focuses on the distance between positive pairs and negative pairs containing one same probe image.
The second term considers the distance between positive pairs and negative pairs which contain different probe images.
With the second constraint, an inter-class distance is supposed to be larger than an intra-class distance.
In\cite{chen2017beyond}, the margin $\beta$ is set to be smaller than the margin $\alpha$ to achieve a relatively weeker constraint, so the second term does not play the leading role.

However, we can well combine these two terms into one and extend \eqref{quadloss} to:
\begin{equation}\label{quadloss2}
\begin{split}
L_{quad^\prime} =  \frac{1}{N}\sum^N \left( \left \| f_A - f_{A^\prime} \right \|_2 - \left \| f_C - f_B \right \|_2 + \alpha \right)_{+}
\end{split}
\end{equation}
where $C$ can share the same identity with $A$ or not.

A direct application of the loss given in \eqref{quadloss2} does not achieve good performance. 
The reason is that the possible number of quadruplets grows rapidly as the dataset gets lager.
The number of all the pairs generated from the quadruplets increases accordingly.
Most of the samples are relatively easy, especially for the negative pairs, the number of which is squarely larger than that of positive ones.
Although a margin is set to restrict the distance between positive and negative pairs, most samples are still too easy to the network, causing the ``precious" hard samples overwhelmed and limiting the model performance.
In order to relieve this, we apply hard sample mining as in \cite{hermans2017defense}.
Triplet loss with hard sample mining computes a batch of samples together.
In each batch, it contains different identities, each of which have the same number of samples.
For each sample, it picks the most dissimilar sample with the same identity and the most similar sample with a different identity to get a triplet. 
In \cite{hermans2017defense}, the triplet loss with hard sample mining is formulated as following:
\begin{equation}\label{trplosshard}
\begin{aligned} 
L_{trihard} = \frac{1}{N}\sum_{A\in batch}  \Bigg ( & \overbrace{\max_{A\prime}(\left \| f_A - f_{A^\prime} \right \|_2)}^{\text{hard positive pair}} \\
 &-\overbrace{\min_{B}(\left \| f_A - f_B \right \|_2)}^{\text{hard negative pair}}+ \alpha \Bigg)_{+}\\
 \end{aligned}
\end{equation}
where $N$ is the batch size.
With hard sample mining, easy samples are filtered and thus improving the robustness of the model.


\subsubsection{Margin sample mining loss}
We apply a new hard example mining strategy for \eqref{quadloss2} named margin smaple mining loss (MSML).
It picks the most dissimilar positive pairs and the most similar negative pair in the whole batch, as:
\begin{equation}\label{edgeminingloss}
L_{eml} = \Bigg( \overbrace{\max_{A,A\prime}(\left \| f_A - f_{A^\prime} \right \|_2)}^{\text{hardest positive pair}} -\overbrace{\min_{C,B}(\left \| f_C - f_B \right \|_2)}^{\text{hardest negative pair}}+ \alpha \Bigg)_{+}
\end{equation}
where $C$ and $B$ can share the same identity with A or not.
HardNet \cite{mishchukmrm17} also proposed a loss that maximizes the distance between the closest positive and close negative patch in a batch and shows great performance in some other tasks.

\begin{figure}[b]
\centering
\subfigure[Relative distance]{
    \label{fig:exlossexampla}
    \includegraphics[width=.48\linewidth]{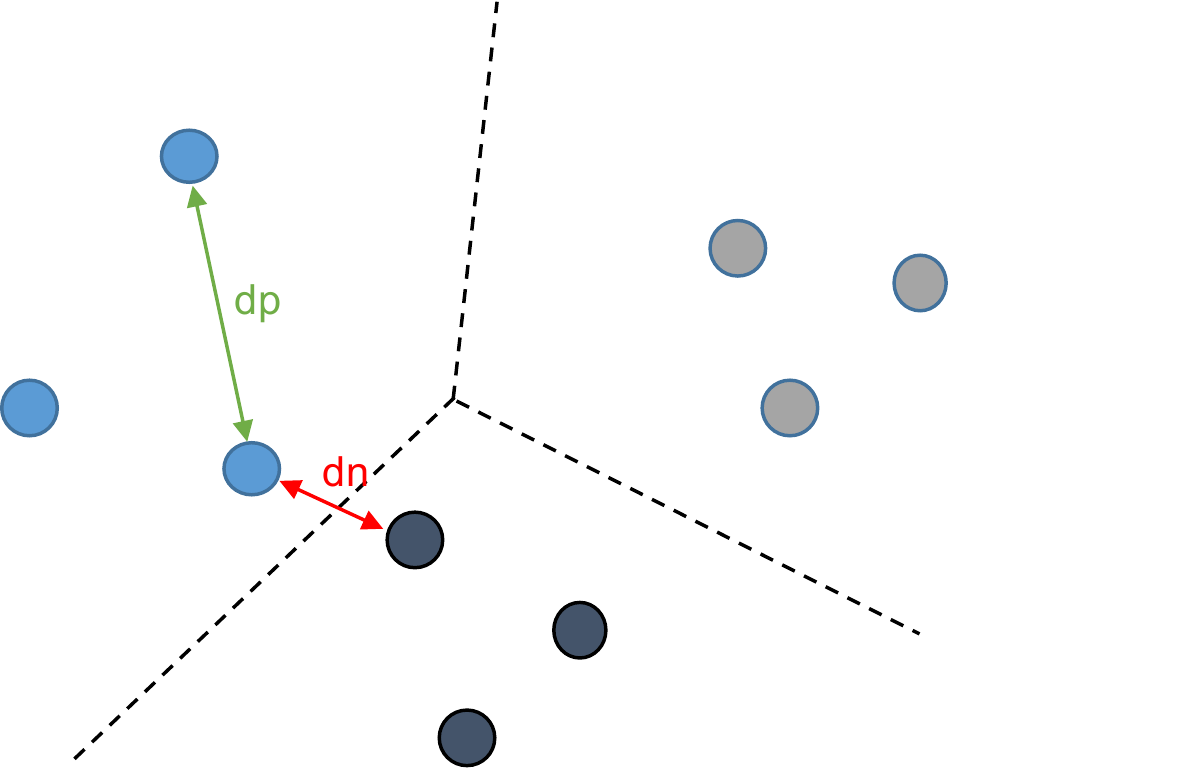}}
\subfigure[Absolute distance]{
    \label{fig:exlossexamplb}
    \includegraphics[width=.48\linewidth]{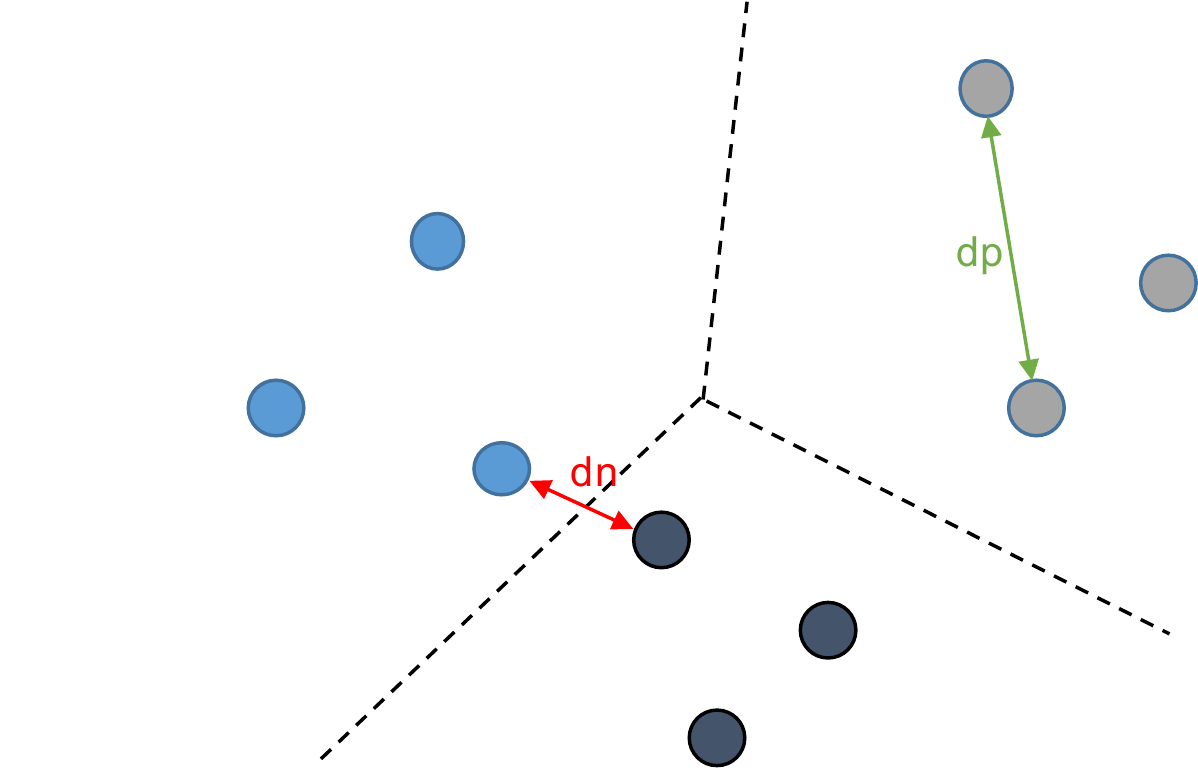}}
\caption{Two examples of edge mining samples.}
\label{fig:exloss}
\end{figure}

As shown in Figure \ref{fig:exloss}, the connections are extremely sparse, only two pairs in a batch participating in training phase.
There are two examples in our margin smaple mining loss.
In Figure \ref{fig:exlossexampla}, the positive pair and the negative pair have one common identity, which considers the relative distance.
It covers the samples that the triplet loss (or with hard sample mining) can get.
And in Figure \ref{fig:exlossexamplb}, the positive pair and the negative pair do not have any common identities, which considers the absolute distance.
Therefore, it can cover the second term of quadruplet loss.
It seems that we waste a lot of training data.
But the two chosen pairs are determined by all the data of one batch.
With the loss reducing, not only the two chosen pairs, but the distances of most positive pairs and negative pairs will get larger.
In addition, we randomly sample the training data in each batch, which allows the pairs diversity as training epoch grows.

In \eqref{edgeminingloss}, the first term is the upper bound of the distance of all positive pairs, and the second term is the lower bound of the distance of all negative pairs in a batch.
Different from other metric learning losses, which push away positive pairs and negative pairs by each sample, our MSML push away the bounds of two sets in a batch.
With training epoch growing, there is a sharp demarcation between positive pairs and negative pairs in feature embedding space.
We think it is a useful characteristic for some special tasks.

In summary, compared with other metric learning losses, our MSML has following advantages.
First, MSML not only considers the relative distances between positive and negative pairs containing the same probe sample, but also considers absolute distances between positive and negative pairs from different probe samples.
Second, it inherits the advantage of hard sample mining and other approaches. 
And we extend it to edge mining, which leads to a better performance.
Finally, we think our MSML is easy to implement and combine with other methods.






\section{Experiments}
\label{experiments}
We first conduct two sets of experiments:
1) to compare different networks on person ReID tasks;
2) to evaluate the performance of different losses.
Then we compare the proposed approach with other state-of-the-art methods.
Note that train a single model using all datasets as \cite{xiao2016learning,zhao2017spindle}.

\subsection{Datasets}
We use public datasets including CUHK03 \cite{Li2014DeepReID}, CUHK-SYSU \cite{xiao2016end}, Market1501 \cite{zheng2015scalable} and MARS \cite{zheng2016mars} in our experiments.

\textbf{CUHH03} contains 14,097 images of 1,467 identities. 
It provides the bounding boxes detected from deformable part models (DPM) and manually labeling. 
In this paper, we evaluate our method on the labeled set. 
Following the evaluation procedure in \cite{Li2014DeepReID}, we randomly pick 100 identities for testing. 
Since we train one single model for all benchmarks, it is a bit different from the standard procedure, which splits the dataset randomly for 20 times.
We only split the dataset once for training and testing.

\begin{table*}[htb]\small 
  \centering
  	\caption{\label{Table2}Comparison of different methods. Cls stands for classfication, Tri stands for triplet loss \cite{schroff2015facenet}, TriHard stands for triplet loss with hard sample mining \cite{hermans2017defense}, Quad stands for quadruplet loss \cite{chen2017beyond} and MSML stands for our margin smaple mining loss. We combine metric learning loss above with classification loss.}
  \begin{tabular}{c|c|ccc|ccc|ccc|ccc}
    \hline
     				& 			& \multicolumn{3}{c|}{Market1501}	& \multicolumn{3}{c|}{MARS}	& \multicolumn{3}{c|}{CUHK-SYSU}	& \multicolumn{3}{c}{CUHK03}	\\
    Base model		& Methods 	& mAP 	& r = 1	& r = 5		& mAP 	& r = 1	&r = 5	& mAP 	& r = 1	&r = 5		& r=1 	&r=5		& r = 10	\\
    \hline
    \hline
     				& Cls 		& 41.3	&65.8	&83.5		& 43.3	&59.3	&75.2	& 70.7	&75.0	&88.1		& 51.2	&72.6	&81.8	\\
     				& Tri 			& 54.8	&75.9	&89.6		& 62.1	&76.1	&89.6	& 82.6	&85.1	&94.1		& 73.0	&92.0	&96.0	\\
    Resnet50		& Quad	 	& 61.1	&80.0	&91.8		& 62.1	&74.9	&88.9	&  85.6	& 87.8	& 95.7& 79.1	&95.3	&97.9	\\
    				& TriHard	 	& 68.0	&83.8	&93.1	&71.3&82.5&92.1& 82.4	&85.1	&94.7		&  79.5	&95.0	& 98.0	\\
    				& MSML	 	&\textbf{69.6}& \textbf{85.2}& \textbf{93.7}	& \textbf{72.0}& \textbf{83.0} & \textbf{92.6}& \textbf{87.2}	& \textbf{89.3}	&\textbf{96.4}		& \textbf{84.0}	&\textbf{96.7}	&\textbf{98.2}	\\
   \hline
        				& Cls 		& 40.7	&66.3	&84.1		& 45.0	&62.6	&77.9	& 74.2	&78.2	&89.7		& 50.5	&68.8	&77.4	\\
    		 		& Tri	 		& 57.9	&78.3	&91.8		& 55.5	&70.7	&85.2	& 87.7	&89.7	&96.6		& 76.9	&93.7	&97.2	\\
    Inception-v2 		& Quad	 	& 66.2	&83.9	&93.6		& 65.3	&77.8	&89.9	& 88.3	&90.2	&96.6		& 81.9	&96.1	&98.3	\\
    			       	& TriHard	 	& 73.2	&86.8	&\textbf{95.4}	& 74.3	&84.1	&93.5	& 83.5	&86.1	&95.2		& 85.5	&97.2	&98.7	\\    		 		
    				& MSML	 	& \textbf{73.4}&\textbf{87.7}&95.2	& \textbf{74.6}&\textbf{84.2}&\textbf{95.1}& \textbf{88.4}&\textbf{90.4}&\textbf{96.8}	& \textbf{86.3}	&\textbf{97.5}&\textbf{98.7}\\
  \hline
       				& Cls 		& 46.5	&70.8	&87.0		& 48.0	&63.8 	&80.2 	& 74.2	&78.2	&89.7		& 57.2	&77.7	&85.6	\\
    		 		& Tri	 		& 69.2	&86.2	&94.7		& 68.2	&79.5	&91.7	& \textbf{89.6}	&\textbf{91.4}&97.0		& 82.0	&96.3	&98.4	\\
    Resnet50-X 		& Quad	 	& 64.8	&83.3	&93.8		& 63.6	&77.7	&89.4	& 87.3	&89.6	&96.2		& 80.7	&94.9	&97.9	\\
    			       	& TriHard	 	& 71.6	&86.9	&94.7		& 69.9	&82.5	&92.4	& 86.4 	&88.8	&96.3		& 82.8	&96.1	&98.1	\\    		 		
    				& MSML	 	& \textbf{76.7}&\textbf{88.9}&\textbf{95.6}& \textbf{72.0}&\textbf{83.4}&\textbf{93.3}& \textbf{89.6}&90.9&\textbf{97.4}	& \textbf{87.5}	&\textbf{97.7}&\textbf{98.9}\\
  \hline
  \end{tabular}
\end{table*}

\textbf{CUHK-SYSU} is a large scale benchmark for person search, containing 18,184 images and 8,432 identities.
The dataset is close to real world application scenarios for images are cropped from whole images.
The training set contains 11,206 images of 5,532 query persons while the test set contains 6,978 images of 2,900 persons.

\textbf{Market1501} contains more than 25,000 images of 1,501 labeled persons of 6 camera views. 
There are 751 identities in the training set and 750 identities in the testing set.
In average, each identity contains 17.2 images with different appearances. 
All images are detected by the DPM detector and thus include 2,793 false alarms to mimic the real scenario.
\textbf{MARS} (Motion Analysis and Re-identification Set) dataset is an extenstion verion of the Market1501 dataset. 
It is a large scale video based person ReID dataset. 
Since all bounding boxes and tracklets are generated automatically, it contains distractors and each identity may have more than one tracklets.
In total, MARS has 20,478 tracklets of 1,261 identities of 6 camera views.

We evaluate our method with rank-1, 5, 10 accuracy and mean average precision (mAP), where the rank-\emph{i} accuracy is the mean accuracy that images of the same identity appear in top-\emph{i}. 
For each query, we calculate the average precision (AP).
And the mean of the average precision (mAP) shows the performance in another dimension.
\subsection{Implementation Details}
Each image is resized into $224\times224$ pixels and conducted with data augmentaion.
The augmentation includes randomly horizontal flipping, shifting, zooming and bluring.
The base models (Resnet50, Inception-v2, Resnet50-Xception (Resnet50-X)) are pre-trained from ImageNet dataset.
The final feature dimensions of Resnet50, Inception-v2 and Resnet50-X are transformed to $1024$ through a fully-connected layer. 
The margin of triplet loss is set to $\alpha = 0.3$ and the margins of the quadruplet loss is set to $\alpha = 0.3$ and $\beta = 0.2$.
The margin of triplet loss with hard mining and our loss with edge mining are also set to $\alpha = 0.3$.
Adam optimizer is used and the inital learning rate is set to $10^{-3}$ in the first $50$ epoches.
Learning rate decreases to $10^{-4}$ in the next $150$ epoches and $10^{-5}$ until convergence.
And the batch size is set to $128$.

We use Resnet50, Inception-v2 and Resnet50-X as base model respectively with different loss functions.
There are several contrast experiments and the results are shown in Table \ref{Table2}.

\subsection{Results analysis of Different Losses}

\begin{figure}[htb]
\centering
\subfigure[TriHard]{
    \label{fig:triplethardsphere}
    \includegraphics[width=0.48\linewidth]{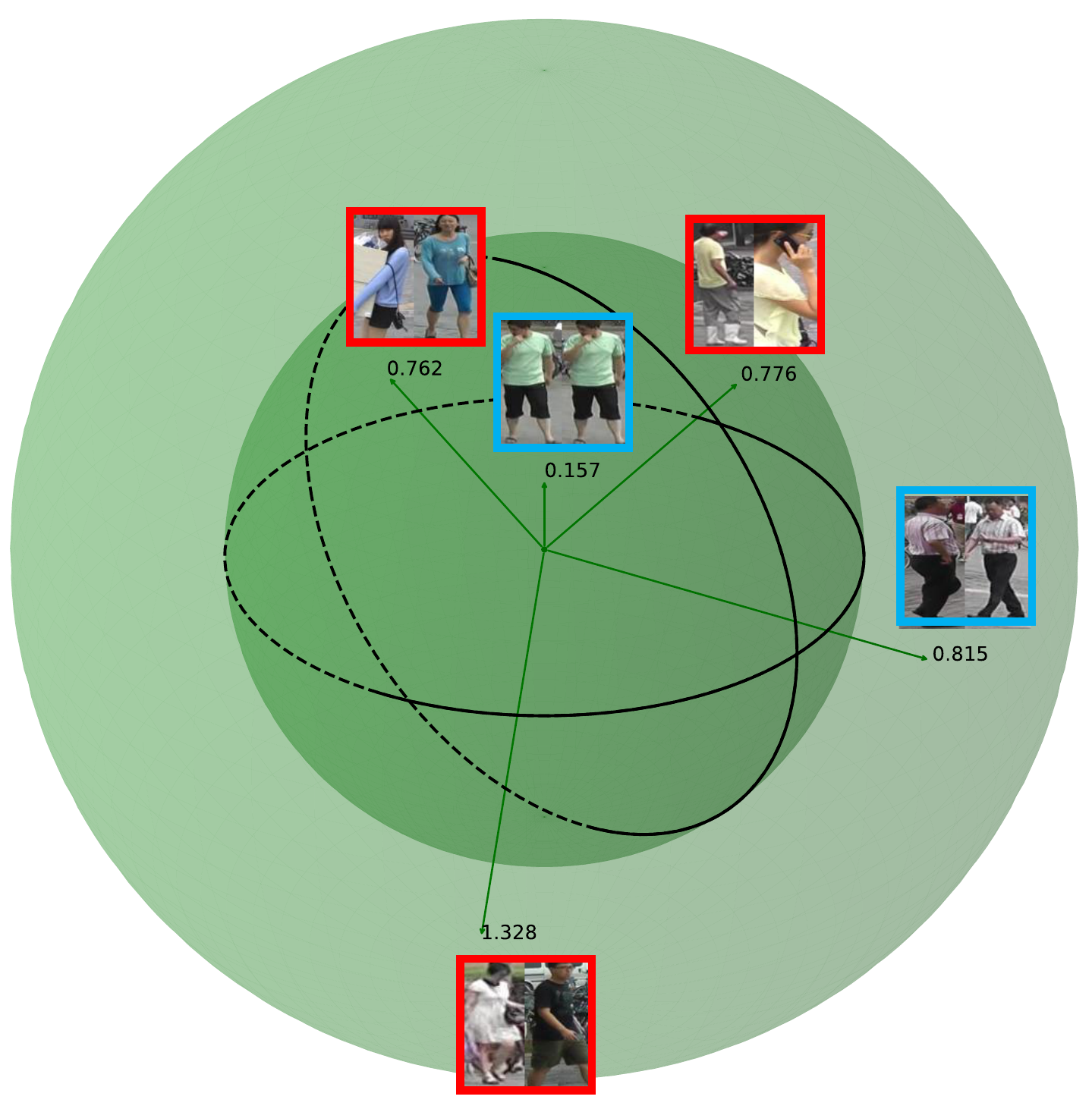}}
\subfigure[MSML]{
    \label{fig:edgeminingsphere}
    \includegraphics[width=0.48\linewidth]{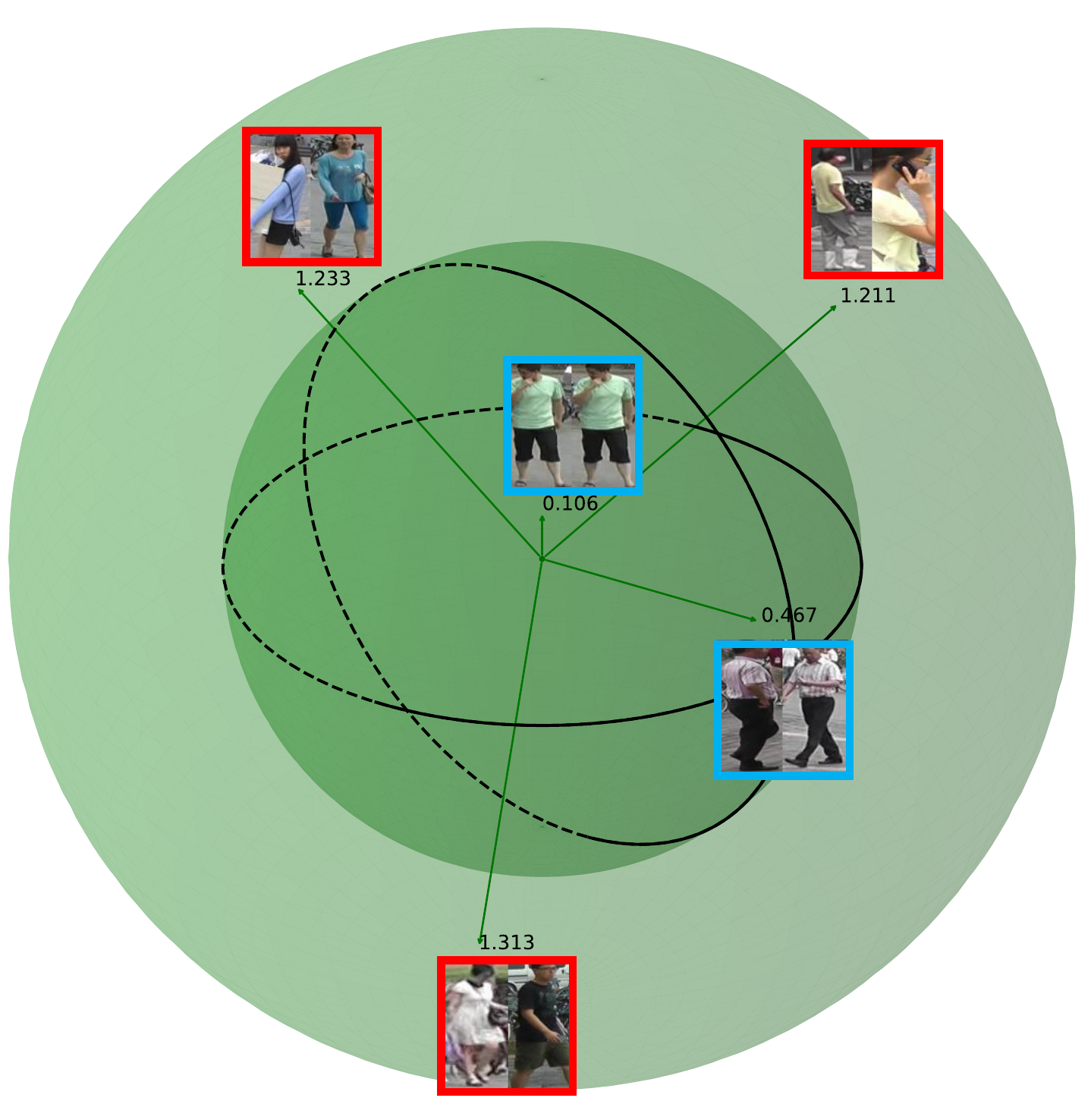}}
\caption{Distance distributions of two different metric learning losses. Blue boxes are positive pairs while red boxes are negative pairs. Note that the direction arrows are only used for viewing.}
\label{fig:1dimspace}
\end{figure}

We conduct experiments with different losses and provide the results to illustrate the effectiveness of our proposed MSML.
They are shown in Table \ref{Table2}.
Cls (classification loss) is the baseline experiment.
Then, we combine different metric learning losses with classification loss.
For Tri (triplet loss), the mAP and rank-1 accuracy increase by approximately $10.0\%$ compared to baseline experiments.
TriHard (triplet loss with hard sample mining) and Quad (quadruplet loss) both have better performance than triplet loss.
TriHard is a little better on Market1501, MARS and CUHK-03 while Quad does better on CUHK-SYSU.
Finally, our MSML gets best accuracy on most experiments datasets for all different base models.

In terms of accuracy, TriHard and MSML can both get high scores.
We further visualize the distance distributions of some randomly chosen image pairs in Figure \ref{fig:1dimspace}.
The numeric values below the image pairs stand for the distances of their features in the embedding space.
As we can see, the distances of negative pairs may be smaller than positive pairs, because TriHard does not focus on absolute distance.
In contrast, our MSML can get a finer metric in feature embedding space.

For Quad and TriHard, some experiments were unable to reach its best accuracy with the same setting. And in Inception-v2 and Resnet-X experiments, they can be even worse than Tri. However, compared with them, our MSML can always have the best performance.

\begin{table}[t]\small
  \centering
      \caption{\label{table_market1501}Comparison on \textbf{Market1501} with single query}
  \begin{tabular}{c|cc}
    	\hline
	Methods  							& mAP	& r=1			\\
	\hline
	\hline
	Temporal \cite{martinel2016temporal}	&22.3	&47.9	\\
	Learning \cite{zhang2016learning}		&35.7	&61.0	\\
	Gated \cite{varior2016gated}			&39.6	&65.9		\\
	Person \cite{chen2017person}			&45.5	&71.8		\\
	Pose \cite{zheng2017pose}			&56.0	&79.3		\\
	Scalable \cite{bai2017scalable}			&68.8	&82.2	\\
	Improving \cite{lin2017improving}		&64.7	&84.3	\\
	In \cite{hermans2017defense}			&69.1	&84.9		\\
	Spindle\cite{zhao2017spindle}			&-		&76.9	\\
	Deep\cite{zhang2017deep}$^*$		&68.8	&87.7	\\
   	\hline
	Our								&\textbf{76.7}	&\textbf{88.9}	\\
  \hline
  \end{tabular}
  \label{market1501}
\end{table}

\begin{table}[t]\small
  \centering
      \caption{\label{table_mars}Comparison on \textbf{MARS} with single query}
  \begin{tabular}{c|cc}
    	\hline
	Methods  		& mAP	& r=1			\\
	\hline
	\hline
	Re-ranking \cite{zhong2017re}			&68.5&73.9 \\
	Learning \cite{zhang2017learning}$^*$ 	&-&55.5\\
	Multi \cite{tesfaye2017multi}$^*$		&-  &68.2	\\
	Mars \cite{zheng2016mars}			& 49.3	&68.3	\\
	In \cite{hermans2017defense}			&67.7&79.8	\\
	Quality \cite{liu2017quality}$^*$			&51.7&73.7	\\
	See \cite{zhousee}					& 50.7	&70.6	\\	
   	\hline
	Our		& \textbf{74.6}	&\textbf{84.2}		\\
  \hline
  \end{tabular}
	\label{mars}
\end{table}

\subsection{Comparison with state-of-the-art methods}
We compare our method with representative ReID methods on several benchmark datasets ($^*$ means it is on ArXiv but not published). The results are shown in Table \ref{table_market1501}, \ref{table_mars}, \ref{table_cuhk03}, \ref{table_cuhksysu}. Methods which applied re-ranking\cite{zhong2017re} skills are not included.

\begin{table}[t]\small
  \centering
      \caption{\label{table_cuhk03}Comparison with existing methods on \textbf{CUHK03}}
  \begin{tabular}{c|ccc}
    	\hline
	Methods  							& r=1	& r=5	&r=10		\\
	\hline
	\hline
	Person \cite{liao2015person} 			& 44.6	& -		&-\\
	Learning \cite{zhang2016learning}		& 62.6	&90.0 	&94.8 \\
	Gated \cite{varior2016gated} 			& 61.8	& -		&-\\
	A \cite{varior2016a} 					& 57.3  	& 80.1	&88.3\\
	In \cite{hermans2017defense} 			& 75.5	&95.2	&\textbf{99.2}\\
	Joint \cite{xiao2017joint}				& 77.5	& -		&-\\
	Deep \cite{geng2016deep}$^*$			& 84.1	& -		&-\\
	Looking \cite{barbosa2017looking}$^*$	& 72.4	&95.2	&95.8\\
	Unlabeled \cite{zheng2017unlabeled}$^*$& 84.6 	& 97.6	&98.9\\
	A \cite{zheng2016discriminatively}$^*$	& 83.4	& 97.1	&98.7\\
	Spindle\cite{zhao2017spindle}			& \textbf{88.5}	&\textbf{97.8}	&98.6\\
   	\hline
	Our								&87.5	&97.7	&98.9	\\
  \hline
  \end{tabular}
  \label{cuhk-sysu}
\end{table}

\begin{table}[t]\small
  \centering
      \caption{\label{table_cuhksysu}Comparison with existing methods on \textbf{CUHK-SYSU}}
  \begin{tabular}{c|cc}
    	\hline
	Methods  						& mAP	& r=1			\\
	\hline
	\hline
	End\cite{xiao2016end}			& 55.7&62.7		\\
	Neural \cite{liu2017neural}$^*$		&77.9&81.2	\\
	Deep \cite{schumann2016deep}$^*$&74.0&76.7 \\
   	\hline
	Our							&\textbf{89.6}&\textbf{90.9} \\
  \hline
  \end{tabular}
\end{table}


\section{Conclusion}
\label{conclusion}
In this paper, we propose a new metric learning loss with hard sample mining named MSML in person re-identification (ReID).
For triplet and quadruplet loss, the positive pairs and negative pairs are randomly sampled.
With hard sample mining, easy samples are filtered and thus improving the robustness of the model.
In our method, we calculate a distance matrix and then choose the maximum distance of positive pairs and the minimum distance of negative pairs to calculate the final loss.
In this way, MSML uses the most dissimilar positive pair and most similar negative pair to train the model.

We use Resnet50, Inception-v2 and Resnet50-X as base models to do some contrast experiments with different metric learning losses.
The results show our MSML gets best performance and learns a finer metric in feature embedding space.
Then, we compare our method with some state-of-the art methods.
On several benchmark datasets, including Market1501, MARS, CUHK-SYSU and CUHK-03, our method shows better performance than most of other methods.

\clearpage
{\small
\bibliographystyle{ieee}
\bibliography{ref}

\begin{thebibliography}{10}\itemsep=-1pt

\bibitem{bai2017scalable}
S.~Bai, X.~Bai, and Q.~Tian.
\newblock Scalable person re-identification on supervised smoothed manifold.
\newblock {\em arXiv preprint arXiv:1703.08359}, 2017.

\bibitem{barbosa2017looking}
I.~B. Barbosa, M.~Cristani, B.~Caputo, A.~Rognhaugen, and T.~Theoharis.
\newblock Looking beyond appearances: Synthetic training data for deep cnns in
  re-identification.
\newblock {\em arXiv preprint arXiv:1701.03153}, 2017.

\bibitem{chen2017beyond}
W.~Chen, X.~Chen, J.~Zhang, and K.~Huang.
\newblock Beyond triplet loss: a deep quadruplet network for person
  re-identification.
\newblock {\em arXiv preprint arXiv:1704.01719}, 2017.

\bibitem{chen2017person}
Y.-C. Chen, X.~Zhu, W.-S. Zheng, and J.-H. Lai.
\newblock Person re-identification by camera correlation aware feature
  augmentation.
\newblock {\em IEEE Transactions on Pattern Analysis and Machine Intelligence},
  2017.

\bibitem{cheng2016person}
D.~Cheng, Y.~Gong, S.~Zhou, J.~Wang, and N.~Zheng.
\newblock Person re-identification by multi-channel parts-based cnn with
  improved triplet loss function.
\newblock In {\em Proceedings of the IEEE Conference on Computer Vision and
  Pattern Recognition}, pages 1335--1344, 2016.

\bibitem{chollet2016xception}
F.~Chollet.
\newblock Xception: Deep learning with depthwise separable convolutions.
\newblock {\em arXiv preprint arXiv:1610.02357}, 2016.

\bibitem{fan2017unsupervised}
H.~Fan, L.~Zheng, and Y.~Yang.
\newblock Unsupervised person re-identification: Clustering and fine-tuning.
\newblock {\em arXiv preprint arXiv:1705.10444}, 2017.

\bibitem{farenzena2010person}
M.~Farenzena, L.~Bazzani, A.~Perina, V.~Murino, and M.~Cristani.
\newblock Person re-identification by symmetry-driven accumulation of local
  features.
\newblock In {\em Computer Vision and Pattern Recognition (CVPR), 2010 IEEE
  Conference on}, pages 2360--2367. IEEE, 2010.

\bibitem{geng2016deep}
M.~Geng, Y.~Wang, T.~Xiang, and Y.~Tian.
\newblock Deep transfer learning for person re-identification.
\newblock {\em arXiv preprint arXiv:1611.05244}, 2016.

\bibitem{hamdoun2008person}
O.~Hamdoun, F.~Moutarde, B.~Stanciulescu, and B.~Steux.
\newblock Person re-identification in multi-camera system by signature based on
  interest point descriptors collected on short video sequences.
\newblock In {\em Distributed Smart Cameras, 2008. ICDSC 2008. Second ACM/IEEE
  International Conference on}, pages 1--6. IEEE, 2008.

\bibitem{he2016deep}
K.~He, X.~Zhang, S.~Ren, and J.~Sun.
\newblock Deep residual learning for image recognition.
\newblock In {\em Proceedings of the IEEE conference on computer vision and
  pattern recognition}, pages 770--778, 2016.

\bibitem{hermans2017defense}
A.~Hermans, L.~Beyer, and B.~Leibe.
\newblock In defense of the triplet loss for person re-identification.
\newblock {\em arXiv preprint arXiv:1703.07737}, 2017.

\bibitem{huang2016densely}
G.~Huang, Z.~Liu, K.~Q. Weinberger, and L.~van~der Maaten.
\newblock Densely connected convolutional networks.
\newblock {\em arXiv preprint arXiv:1608.06993}, 2016.

\bibitem{koestinger2012large}
M.~Koestinger, M.~Hirzer, P.~Wohlhart, P.~M. Roth, and H.~Bischof.
\newblock Large scale metric learning from equivalence constraints.
\newblock In {\em Computer Vision and Pattern Recognition (CVPR), 2012 IEEE
  Conference on}, pages 2288--2295. IEEE, 2012.

\bibitem{krizhevsky2012imagenet}
A.~Krizhevsky, I.~Sutskever, and G.~E. Hinton.
\newblock Imagenet classification with deep convolutional neural networks.
\newblock In {\em Advances in neural information processing systems}, pages
  1097--1105, 2012.

\bibitem{li2017person}
S.~Li, T.~Xiao, H.~Li, B.~Zhou, D.~Yue, and X.~Wang.
\newblock Person search with natural language description.
\newblock {\em arXiv preprint arXiv:1702.05729}, 2017.

\bibitem{Li2014DeepReID}
W.~Li, R.~Zhao, T.~Xiao, and X.~Wang.
\newblock Deepreid: Deep filter pairing neural network for person
  re-identification.
\newblock pages 152--159, 2014.

\bibitem{liao2015person}
S.~Liao, Y.~Hu, X.~Zhu, and S.~Z. Li.
\newblock Person re-identification by local maximal occurrence representation
  and metric learning.
\newblock In {\em Proceedings of the IEEE Conference on Computer Vision and
  Pattern Recognition}, pages 2197--2206, 2015.

\bibitem{lin2017improving}
Y.~Lin, L.~Zheng, Z.~Zheng, Y.~Wu, and Y.~Yang.
\newblock Improving person re-identification by attribute and identity
  learning.
\newblock {\em arXiv preprint arXiv:1703.07220}, 2017.

\bibitem{liu2017neural}
H.~Liu, J.~Feng, Z.~Jie, K.~Jayashree, B.~Zhao, M.~Qi, J.~Jiang, and S.~Yan.
\newblock Neural person search machines.
\newblock {\em arXiv preprint arXiv:1707.06777}, 2017.

\bibitem{liu2017end}
H.~Liu, J.~Feng, M.~Qi, J.~Jiang, and S.~Yan.
\newblock End-to-end comparative attention networks for person
  re-identification.
\newblock {\em IEEE Transactions on Image Processing}, 2017.

\bibitem{liu2017video}
H.~Liu, Z.~Jie, K.~Jayashree, M.~Qi, J.~Jiang, S.~Yan, and J.~Feng.
\newblock Video-based person re-identification with accumulative motion
  context.
\newblock {\em arXiv preprint arXiv:1701.00193}, 2017.

\bibitem{liu2017quality}
Y.~Liu, J.~Yan, and W.~Ouyang.
\newblock Quality aware network for set to set recognition.
\newblock {\em arXiv preprint arXiv:1704.03373}, 2017.

\bibitem{ma2017person}
X.~Ma, X.~Zhu, S.~Gong, X.~Xie, J.~Hu, K.-M. Lam, and Y.~Zhong.
\newblock Person re-identification by unsupervised video matching.
\newblock {\em Pattern Recognition}, 65:197--210, 2017.

\bibitem{martinel2016temporal}
N.~Martinel, A.~Das, C.~Micheloni, and A.~K. Roy-Chowdhury.
\newblock Temporal model adaptation for person re-identification.
\newblock In {\em European Conference on Computer Vision}, pages 858--877.
  Springer, 2016.

\bibitem{matsukawa2016person}
T.~Matsukawa and E.~Suzuki.
\newblock Person re-identification using cnn features learned from combination
  of attributes.
\newblock In {\em Pattern Recognition (ICPR), 2016 23rd International
  Conference on}, pages 2428--2433. IEEE, 2016.

\bibitem{mclaughlin2016recurrent}
N.~McLaughlin, J.~Martinez~del Rincon, and P.~Miller.
\newblock Recurrent convolutional network for video-based person
  re-identification.
\newblock In {\em Proceedings of the IEEE Conference on Computer Vision and
  Pattern Recognition}, pages 1325--1334, 2016.

\bibitem{mishchukmrm17}
A.~Mishchuk, D.~Mishkin, F.~Radenovic, and J.~Matas.
\newblock Working hard to know your neighbor's margins: Local descriptor
  learning loss.
\newblock {\em CoRR}, abs/1705.10872, 2017.

\bibitem{peng2016unsupervised}
P.~Peng, T.~Xiang, Y.~Wang, M.~Pontil, S.~Gong, T.~Huang, and Y.~Tian.
\newblock Unsupervised cross-dataset transfer learning for person
  re-identification.
\newblock In {\em Proceedings of the IEEE Conference on Computer Vision and
  Pattern Recognition}, pages 1306--1315, 2016.

\bibitem{radenovic2016cnn}
F.~Radenovi{\'c}, G.~Tolias, and O.~Chum.
\newblock Cnn image retrieval learns from bow: Unsupervised fine-tuning with
  hard examples.
\newblock In {\em European Conference on Computer Vision}, pages 3--20.
  Springer, 2016.

\bibitem{rosasco2004loss}
L.~Rosasco, E.~De~Vito, A.~Caponnetto, M.~Piana, and A.~Verri.
\newblock Are loss functions all the same?
\newblock {\em Neural Computation}, 16(5):1063--1076, 2004.

\bibitem{russakovsky2015imagenet}
O.~Russakovsky, J.~Deng, H.~Su, J.~Krause, S.~Satheesh, S.~Ma, Z.~Huang,
  A.~Karpathy, A.~Khosla, M.~Bernstein, et~al.
\newblock Imagenet large scale visual recognition challenge.
\newblock {\em International Journal of Computer Vision}, 115(3):211--252,
  2015.

\bibitem{schroff2015facenet}
F.~Schroff, D.~Kalenichenko, and J.~Philbin.
\newblock Facenet: A unified embedding for face recognition and clustering.
\newblock In {\em Proceedings of the IEEE Conference on Computer Vision and
  Pattern Recognition}, pages 815--823, 2015.

\bibitem{schumann2016deep}
A.~Schumann, S.~Gong, and T.~Schuchert.
\newblock Deep learning prototype domains for person re-identification.
\newblock {\em arXiv preprint arXiv:1610.05047}, 2016.

\bibitem{zheng2016mars}
Springer.
\newblock {\em MARS: A Video Benchmark for Large-Scale Person
  Re-identification}, 2016.

\bibitem{szegedy2015going}
C.~Szegedy, W.~Liu, Y.~Jia, P.~Sermanet, S.~Reed, D.~Anguelov, D.~Erhan,
  V.~Vanhoucke, and A.~Rabinovich.
\newblock Going deeper with convolutions.
\newblock In {\em Proceedings of the IEEE conference on computer vision and
  pattern recognition}, pages 1--9, 2015.

\bibitem{tesfaye2017multi}
Y.~T. Tesfaye, E.~Zemene, A.~Prati, M.~Pelillo, and M.~Shah.
\newblock Multi-target tracking in multiple non-overlapping cameras using
  constrained dominant sets.
\newblock {\em arXiv preprint arXiv:1706.06196}, 2017.

\bibitem{varior2016gated}
R.~R. Varior, M.~Haloi, and G.~Wang.
\newblock Gated siamese convolutional neural network architecture for human
  re-identification.
\newblock In {\em European Conference on Computer Vision}, pages 791--808.
  Springer, 2016.

\bibitem{varior2016a}
R.~R. Varior, B.~Shuai, J.~Lu, D.~Xu, and G.~Wang.
\newblock A siamese long short-term memory architecture for human
  re-identification.
\newblock In {\em European Conference on Computer Vision}, pages 135--153,
  2016.

\bibitem{wang2017deep}
J.~Wang, S.~Zhou, J.~Wang, and Q.~Hou.
\newblock Deep ranking model by large adaptive margin learning for person
  re-identification.
\newblock {\em arXiv preprint arXiv:1707.00409}, 2017.

\bibitem{wang2016person}
T.~Wang, S.~Gong, X.~Zhu, and S.~Wang.
\newblock Person re-identification by discriminative selection in video
  ranking.
\newblock {\em IEEE transactions on pattern analysis and machine intelligence},
  38(12):2501--2514, 2016.

\bibitem{xiao2016learning}
T.~Xiao, H.~Li, W.~Ouyang, and X.~Wang.
\newblock Learning deep feature representations with domain guided dropout for
  person re-identification.
\newblock In {\em Proceedings of the IEEE Conference on Computer Vision and
  Pattern Recognition}, pages 1249--1258, 2016.

\bibitem{xiao2016end}
T.~Xiao, S.~Li, B.~Wang, L.~Lin, and X.~Wang.
\newblock End-to-end deep learning for person search.
\newblock {\em arXiv preprint arXiv:1604.01850}, 2016.

\bibitem{xiao2017joint}
T.~Xiao, S.~Li, B.~Wang, L.~Lin, and X.~Wang.
\newblock Joint detection and identification feature learning for person
  search.
\newblock In {\em Proc. CVPR}, 2017.

\bibitem{xie2016aggregated}
S.~Xie, R.~Girshick, P.~Doll{\'a}r, Z.~Tu, and K.~He.
\newblock Aggregated residual transformations for deep neural networks.
\newblock {\em arXiv preprint arXiv:1611.05431}, 2016.

\bibitem{you2016top}
J.~You, A.~Wu, X.~Li, and W.-S. Zheng.
\newblock Top-push video-based person re-identification.
\newblock In {\em Proceedings of the IEEE Conference on Computer Vision and
  Pattern Recognition}, pages 1345--1353, 2016.

\bibitem{zhang2017image}
D.~Zhang, W.~Wu, H.~Cheng, R.~Zhang, Z.~Dong, and Z.~Cai.
\newblock Image-to-video person re-identification with temporally memorized
  similarity learning.
\newblock {\em IEEE Transactions on Circuits and Systems for Video Technology},
  2017.

\bibitem{Zhang_2016_CVPR}
L.~Zhang, T.~Xiang, and S.~Gong.
\newblock Learning a discriminative null space for person re-identification.
\newblock In {\em The IEEE Conference on Computer Vision and Pattern
  Recognition (CVPR)}, June 2016.

\bibitem{zhang2016learning}
L.~Zhang, T.~Xiang, and S.~Gong.
\newblock Learning a discriminative null space for person re-identification.
\newblock In {\em Proceedings of the IEEE Conference on Computer Vision and
  Pattern Recognition}, pages 1239--1248, 2016.

\bibitem{zhang2017learning}
W.~Zhang, S.~Hu, and K.~Liu.
\newblock Learning compact appearance representation for video-based person
  re-identification.
\newblock {\em arXiv preprint arXiv:1702.06294}, 2017.

\bibitem{zhang2017shufflenet}
X.~Zhang, X.~Zhou, M.~Lin, and J.~Sun.
\newblock Shufflenet: An extremely efficient convolutional neural network for
  mobile devices.
\newblock {\em arXiv preprint arXiv:1707.01083}, 2017.

\bibitem{zhang2017deep}
Y.~Zhang, T.~Xiang, T.~M. Hospedales, and H.~Lu.
\newblock Deep mutual learning.
\newblock {\em arXiv preprint arXiv:1706.00384}, 2017.

\bibitem{zhao2017spindle}
H.~Zhao, M.~Tian, S.~Sun, J.~Shao, J.~Yan, S.~Yi, X.~Wang, and X.~Tang.
\newblock Spindle net: Person re-identification with human body region guided
  feature decomposition and fusion.
\newblock CVPR, 2017.

\bibitem{zhao2017person}
R.~Zhao, W.~Oyang, and X.~Wang.
\newblock Person re-identification by saliency learning.
\newblock {\em IEEE transactions on pattern analysis and machine intelligence},
  39(2):356--370, 2017.

\bibitem{zheng2017pose}
L.~Zheng, Y.~Huang, H.~Lu, and Y.~Yang.
\newblock Pose invariant embedding for deep person re-identification.
\newblock {\em arXiv preprint arXiv:1701.07732}, 2017.

\bibitem{zheng2015scalable}
L.~Zheng, L.~Shen, L.~Tian, S.~Wang, J.~Wang, and Q.~Tian.
\newblock Scalable person re-identification: A benchmark.
\newblock In {\em Computer Vision, IEEE International Conference}, 2015.

\bibitem{zheng2016person}
L.~Zheng, Y.~Yang, and A.~G. Hauptmann.
\newblock Person re-identification: Past, present and future.
\newblock {\em arXiv preprint arXiv:1610.02984}, 2016.

\bibitem{zheng2016discriminatively}
Z.~Zheng, L.~Zheng, and Y.~Yang.
\newblock A discriminatively learned cnn embedding for person
  re-identification.
\newblock {\em arXiv preprint arXiv:1611.05666}, 2016.

\bibitem{zheng2017unlabeled}
Z.~Zheng, L.~Zheng, and Y.~Yang.
\newblock Unlabeled samples generated by gan improve the person
  re-identification baseline in vitro.
\newblock {\em arXiv preprint arXiv:1701.07717}, 2017.

\bibitem{zhong2017re}
Z.~Zhong, L.~Zheng, D.~Cao, and S.~Li.
\newblock Re-ranking person re-identification with k-reciprocal encoding.
\newblock {\em arXiv preprint arXiv:1701.08398}, 2017.

\bibitem{zhousee}
Z.~Zhou, Y.~Huang, W.~Wang, L.~Wang, and T.~Tan.
\newblock See the forest for the trees: Joint spatial and temporal recurrent
  neural networks for video-based person re-identification.

\end{thebibliography}
}


\end{document}